\documentclass[10pt,twocolumn,letterpaper]{article}

\usepackage{wacv}              

\usepackage{graphicx}
\usepackage{amsmath}
\usepackage{amssymb}
\usepackage{booktabs}

\usepackage[pagebackref,breaklinks,colorlinks]{hyperref}

\usepackage{makecell}
\usepackage{tabularx}
\usepackage{multirow}
\usepackage{xcolor}
\usepackage{caption}

\usepackage[colorinlistoftodos,color=pink]{todonotes} 
\usepackage{xkeyval}
\presetkeys{todonotes}{inline}{}

\usepackage{lipsum} 
\usepackage{pgfplots}

\usepackage[capitalize]{cleveref}
\crefname{section}{Sec.}{Secs.}
\Crefname{section}{Section}{Sections}
\Crefname{table}{Table}{Tables}
\crefname{table}{Tab.}{Tabs.}

\def\wacvPaperID{1629} 
\def\confName{WACV}
\def\confYear{2025}

\begin{document}

\title{Robust Long-Range Perception Against \\ Sensor Misalignment in Autonomous Vehicles} 

\author{Zi-Xiang Xia, Sudeep Fadadu, Yi Shi, Louis Foucard\\
Aurora Innovation, Inc. \\
{\tt\small \{sxia, sfadadu, yishi, flouis\}@aurora.tech}
}
\maketitle


\begin{abstract}
    Advances in machine learning algorithms for sensor fusion have significantly improved the detection and prediction of other road users, thereby enhancing safety. However, even a small angular displacement in the sensor's placement can cause significant degradation in output, especially at long range. In this paper, we demonstrate a simple yet generic and efficient multi-task learning approach that not only detects misalignment between different sensor modalities but is also robust against them for long-range perception.  Along with the amount of misalignment, our method also predicts calibrated uncertainty, which can be useful for filtering and fusing predicted misalignment values over time. In addition, we show that the predicted misalignment parameters can be used for self-correcting input sensor data, further improving the perception performance under sensor misalignment.  
\end{abstract}

\section{Introduction}
\label{sec:intro}
The rapid evolution of machine learning algorithms \cite{ma2020artificial, chib2023recent} has revolutionized the field of autonomous vehicles and Advanced Driver Assistance Systems (ADAS), paving the way for safer and more efficient operation on the road. Deep learning techniques, in particular, have proven invaluable in processing and making sense of the vast amounts of data collected from various sensors, including cameras, LiDAR, and radar \cite{prakash2021multi}\cite{ Fadadu_2022_WACV}\cite{laddha2021mvfusenet}\cite{Gautam_2021_ICCV}.
\begin{figure*}
    \centering
    \includegraphics[width=0.7\linewidth]{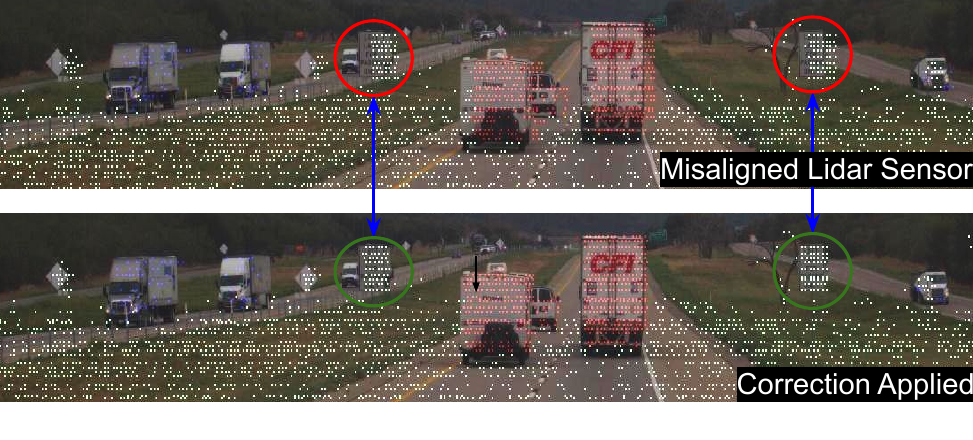}
    \captionsetup{font=small}
    \caption{Demonstration of sensor misalignment correction using the proposed approach. The top image shows an input frame with visible misalignment when projecting the LiDAR data (green points) onto the camera image. In the bottom image, after applying the extrinsic parameter correction predicted by our model, the LiDAR points are well-aligned with the camera data, leading to improved 3D object detection accuracy.}
    \label{fig:cal-corr}
\end{figure*}

The multimodal perception systems combine the strengths of different sensors, providing a more comprehensive and robust understanding of the vehicle's surroundings. For example, camera data offers high-resolution visual information, while LiDAR excels at accurately mapping the 3D environment, and radar is adept at detecting and tracking moving objects even in challenging weather conditions.  

Sensor fusion algorithms and models rely on intrinsic and extrinsic calibration parameters to integrate these different sensors. These parameters define the precise spatial relationships and transformations between the various sensors (such as cameras, LiDAR, and radar) and the vehicle's coordinate frame. These parameters are determined through a process known as calibration. Any deviations from these predetermined extrinsic parameters, caused by factors like vibrations, impacts, or thermal effects, can lead to inconsistencies in the sensor data fusion process. Consequently, this can result in erroneous object detection, localization errors, and ultimately, compromised decision-making capabilities.

For self-driving highway applications, especially heavy-weight trucks, autonomy is highly susceptible and requires extra attention due to the longer stopping distance requirement. A deviation of just 5 milliradians translates to an error of 2.25 meters at 450 meter range, resulting in a complete mismatch of the fused image and LiDAR data, and enough lateral error to place a vehicle on the wrong lane. 

Monitoring the relative positions of onboard sensors can be crucial for safe operations. If the sensors' positions are found to have deviated from their original locations, the planner system of the autonomous vehicle (AV) can take appropriate actions to mitigate risks to acceptable levels, such as stopping on the shoulder or reducing speed. In addition to monitoring, it is essential that the perception models used for detection and tracking tasks are robust against deviations from predetermined calibration parameters. 

In this paper, we propose a system that integrates the misalignment monitoring task with a 3D detection task. We limit our work to only angular displacement in this paper as translational displacement (a few millimeters) has minimal impact on sensor measurements compared to the inherent variability (variance) present in the sensor's output.
 
 To achieve this, we integrate a misalignment prediction head to a base detection network-trunk, and we train the network end to end. This solution leverages joint multi-task learning, offering potential benefits in terms of increased robustness of object detection, accurate misalignment prediction, and efficient computational resource utilization. This approach enables the system to self-correct the alignment errors, enhancing the robustness of long-range detection. Our approach leverages aleatoric (data) uncertainty \cite{kendall2017uncertainties}, which captures the inherent noise and stochasticity in the input sensor data. By explicitly representing this uncertainty in the network's outputs, we can provide well-calibrated confidence estimates for the predicted misalignment values. This enables better-informed decision-making and risk-averse behavior in situations with high uncertainty.

Our contributions are summarized as follows:
  \begin{itemize}
      \item We propose a long-range perception system that is robust against sensor misalignment.
      \item We propose a learning approach with synthetic data augmentation for the sensor misalignment prediction task where ground-truth data is not readily available.
      \item We show that learning misalignment as an auxiliary task serves as a valuable enhancement to object detection models.
      \item Our model predicts well-calibrated uncertainty values for the corresponding misalignment predictions, which can be used further to construct accurate estimate of misalignment over time.
      \item We show that correcting incoming misaligned data with predicted misalignment helps achieve higher detection performance at longer ranges. 

  \end{itemize}

\section{Related Work}

The extrinsic calibration parameters, a 6-DoF transformation between different sensors, are particularly critical in highway driving due to their operational characteristics. High speed vehicles (specially trucks) require significantly longer stopping distances, necessitating the ability to detect even small objects like a construction cone or a pedestrian at a long range, often up to 450 meters. Performing sensor fusion at these ranges requires a very high degree of precision of each sensor's extrinsics parameters.  

\subsection{Calibration} Traditionally, LiDAR and camera offline calibration can be categorized into target-based and target-less methods. Target-based calibration, relies on reference objects such as checkerboards \cite{Zhang2004ExtrinsicCO, Geigercheckboard}, ArUco \cite{ArUco}, and circular grid \cite{circulargrid} to forge a correspondence between LiDAR data and camera images. These methods typically leverage distinctive features like planes and corners to optimize the calibration process, providing high precision within controlled environments. 
While these methods provide high precision in controlled environments, they are limited by the need for such specific settings. In contrast, target-less methods leverage natural environmental features, such as lines (e.g., lines \cite{line1, line2} and edges \cite{edge1, edge2}).  Yuan et al.\cite{Yuan} extracted more accurate and reliable LiDAR edges by using voxel cutting and plane fitting techniques.
However, maintaining a desired level of calibration precision is challenging due to varying environmental and operational factors such as temperature changes, road vibrations, and equipment wear and tear, all of which can introduce calibration errors on an initially well-calibrated system. 

Therefore, online calibration is needed to better handle these conditions on the road. In \cite{onlinecal}, authors monitored the calibration by computing alignment between image edges and projected LiDAR points. An objective function is proposed to model the information of discontinuity from the image and the depth from LiDAR points. Additionally, inspired by the recent success of deep neural networks, online learning-based approaches are also introduced to calibration. RegNet \cite{regnet}, the first deep convolutional neural network for LiDAR-camera calibration marks a significant breakthrough in this field. It is trained on a large amount of miscalibration data to accurately estimate extrinsic parameters. Zhu et al. \cite{Zhu} proposed a semantic segmentation method that first extracts semantic features and combines non-monotonic line search and subgradient ascent to estimate optimal calibration parameters. Sun et al. \cite{Sun} introduced an instance segmentation approach with a cost function that accounts for the matching degree based on the appearance and centroids of key targets in point cloud and image pairs. End-to-end networks, such as  CalibNet \cite{calibnet} incorporate geometric supervision to enhance precision, while LCCNet \cite{lccnet} exploits a cost volume layer for feature matching,  enabling it to capture correlations between different sensors.

Although online calibration methods have been introduced, it's important to acknowledge that perfect calibration might not be achievable and the perception system must also be robust enough to handle unexpected discrepancies in sensor data. For instance, \cite{calibnet} achieves impressive mean error in yaw orientation estimate. However, it is still high enough that LiDAR points corresponding to a pedestrian at a 400-meter range would be projected laterally onto a different lane.  Moreover, these methods are not multi-task methods and require a standalone network alongside a detection network increasing the cost for resource-constrained platforms.  

\subsection{Robust 3D detection} In the field of 3D object detection, especially when working with LiDAR point clouds, data augmentation is crucial to enhance the robustness and accuracy of the detection models. Scaling, rotation, translation, and flipping of the entire point cloud, are common global augmentation methods \cite{global1, global2}. Local augmentations, focusing on individual objects within a point cloud, include techniques like rotation, scaling, and translation of specific ground truth boxes \cite{pattern-aware}. 
Moreover, innovative techniques like PointMixup \cite{pointmixup}, which employs interpolation between point clouds, and InverseAug \cite{deepfusion}, which focuses on reversing augmented features and understanding the correlations between LiDAR and camera features during fusion, have been introduced. Each method offers a unique contribution to enhancing LiDAR point cloud data when leveraging multi-modality fusion, leading to more accurate and dependable 3D object detection. However, none of these methods explicitly perform perturbation of the extrinsic parameters directly.  In our method, we argue that, in addition to the mentioned augmentation and feature interpolation, explicitly perturbing the relative transformation can provide enhanced robustness against sensor misalignment.

\section{Proposed Approach}

\begin{figure}
    \centering
    \includegraphics[width=1\columnwidth]{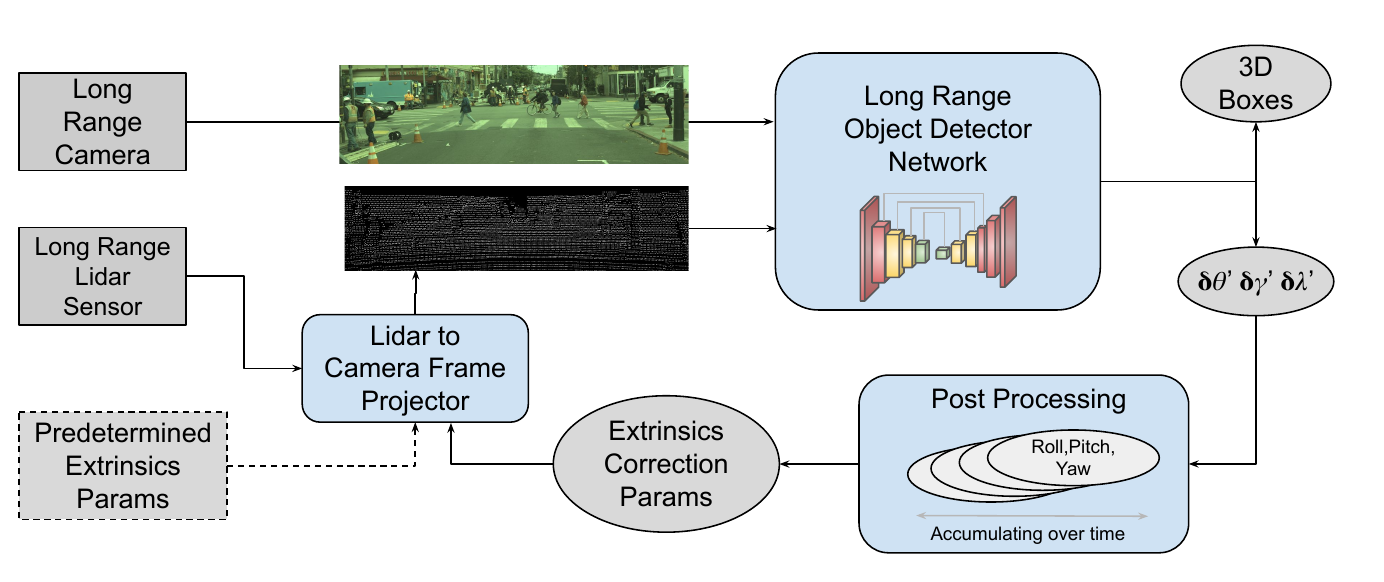}
    \caption{Proposed system architecture for robust long-range perception against sensor misalignment. In the figure, rectangles indicate the system inputs, while circles denote the outputs.}
    \label{fig:sys}
\end{figure}

Our long-range detection system is composed of three phases as described in Figure \ref{fig:sys}. We first start with transforming the incoming LiDAR point-cloud data from LiDAR's sensor frame of reference into the camera's frame of reference using the pre-determined extrinsic parameters. Next, we run inference with our multi-task detector and misalignment predictor on the data as described in section \ref{sec:network}. The estimated misalignment parameters, along with predicted uncertainty are fused over a 5-second window. Finally, the fused misalignment values from history are used in the initial pre-processing phase for correcting transformed points in the camera's frame of current reference.  

In the following sections, we describe each phase in more detail, including model architecture, loss, training details, and the self-correction process.

\subsection{Phase 1: Pre-Processing}
In this phase, we prepare a depth image from the LiDAR point-cloud data. Given extrinsic $T$ and intrinsic $K$, we can generate the depth image by projecting LiDAR points  $P_i$ = $[$$X_i$  $Y_i$  $Z_i ]$ $\in$ $\mathbb{R}^3$ to image pixel coordinate $p_i = [u_i$  $v_i] \in \mathbb{R}^2$. The pixel values are set to $Z_i$ where the LiDAR point is projected. For the remaining pixels, the value is set to zero.


\subsection{Phase 2: Inference}
\label{sec:network}
\subsubsection{Misalignment Estimation} 
In the event of pure rotational displacement of LiDAR sensor, the coordinates of the point-cloud shift accordingly. For rotational displacement of $ \Delta r$ = $[\theta^{roll}, \theta^{pitch}, \theta^{yaw}]$, the projected pixel coordinates in depth image $I_{depth}$, will be shifted
to $p'_i = [u'_i$  $v'_i] \in \mathbb{R}^2$.
 compared to the corresponding pixel locations $p_i$ in the RGB image $I_{rgb}$. 
Since the relative translation component in $T$ is negligible compared to the long ranges we are interested in, we approximate the problem as one of pure rotation with respect to the camera's frame of reference. With this approximation, values of $p'_i$ can be obtained from $p_i$ using a Homography matrix $H$.  
 
\begin{equation}
\label{eq:homography}
\begin{aligned}
\hat{p}'_i & \approx & H \cdot \hat{p}_i \\
     & \approx & K \cdot \Delta R \cdot K^{-1} \cdot \hat{p}_i
\end{aligned}
\end{equation}

where $\hat{p}_i$ and $\hat{p}'_i$ represent the homogeneous coordinates of $p_i$ and $p'_i$, $\Delta R$ is the rotation matrix representation of relative rotation $\Delta r$ between $I_{depth}$ and $I_{rgb}$.

Throughout the research, the intrinsic matrix $K$ is known and fixed. To further simplify the equation, We also assume that the $\Delta r$ is sufficiently small, and small angle approximation \cite{small-angle} can be applied. With $\tilde{p}_i = K^{-1}p_i = [\tilde{x_i}, \tilde{y_i}, \tilde{z_i}]$ and $\tilde{p}'_i = K^{-1}p'_i = [\tilde{x}'_i, \tilde{y}'_i, \tilde{z}'_i]$, the equation \ref{eq:homography} can be rewritten as,

\begin{equation}
\label{eq:homo_Approx}
\begin{aligned}
\begin{bmatrix}
\tilde{x}'_i \\ \tilde{y}'_i \\ \tilde{z}'_i
\end{bmatrix} & \approx & \begin{bmatrix}
1 & -\theta^{yaw} & \theta^{pitch} \\
\theta^{yaw} & 1 & -\theta^{roll} \\
-\theta^{pitch} & \theta^{roll} & 1
\end{bmatrix} \cdot \begin{bmatrix}
\tilde{x}_i \\ \tilde{y}_i \\ \tilde{z}_i
\end{bmatrix} \\
& \approx &  \begin{bmatrix}
\tilde{x}_i -  \theta^{yaw}  \tilde{y}_i + \theta^{pitch}  \tilde{z}_i \\ 
\theta^{yaw} \tilde{x}_i + \tilde{y}_i - \theta^{roll} \tilde{z}_i \\ 
-\theta^{pitch} \tilde{x}_i + \theta^{roll} \tilde{y}_i + \tilde{z}_i
\end{bmatrix}
\end{aligned}
\end{equation}

Given the point location correspondences $\tilde{p}_i$ and $\tilde{p'}_i$, we can set up a linear system $Ax = 0$, where $x = [\theta^{roll}$ $\theta^{pitch}$ $ \theta^{yaw} $ $ 1]^T$. 

\begin{equation}
\begin{bmatrix}
0 & \tilde{z}_1 & -\tilde{y}_1 & \tilde{x}_1 - \tilde{x}'_1 \\
-\tilde{z}_1 & 0 & \tilde{x}_1 & \tilde{y}_1 - \tilde{y}'_1\\
\tilde{y}_1 & -\tilde{x}_1 & 0 & \tilde{z}_1 - \tilde{z}'_1\\

\vdots & \vdots & \vdots & \vdots \\
0 & \tilde{z}_n & -\tilde{y}_n & \tilde{x}_n - \tilde{x}'_n \\
-\tilde{z}_n & 0 & \tilde{x}_n & \tilde{y}_n - \tilde{y}'_n\\
\tilde{y}_n & -\tilde{x}_n & 0 & \tilde{z}_n - \tilde{z}'_n\\
\end{bmatrix}
\begin{bmatrix}
\theta^{roll} \\ \theta^{pitch} \\ \theta^{yaw} \\ 1
\end{bmatrix} =
\begin{bmatrix}
0 \\
0 \\
0 \\
\vdots \\
0 \\
0 \\
0
\end{bmatrix}
\end{equation}

However, obtaining correspondences between $I_{rgb}$ and $I_{depth}$ is a non-trivial task. To address this, we employ multi-task learning to directly regress $\Delta r$ in a similar fashion as demonstrated in \cite{poursaeed2018deep} for fundamental matrix estimation without known correspondences.  For ground-truth $\Delta r$, we perturb the LiDAR points in the camera image frame of reference and train the network to estimate the amount of perturbation. Also note that throughout this work, we assume that the intrinsic matrix does not change significantly between the training and testing sets which makes the matrix $A$ dependent only on the input RGB and depth images.
 
\subsubsection{Network Architecture}
To demonstrate the effectiveness of our method, we use CenterNet \cite{centernet} as our base model. 
CenterNet consists of a convolutional backbone network for feature extraction and decoding heads to output heatmaps and offsets. In this work, we adopt a similar structure, employing VoVNetV2\cite{vovnetv2} as the CNN backbone. Initially, the image is processed through a stem network consisting of two fully convolutional layers with 32 and 64 dimensions, and kernel sizes of 7 × 7 and 3 × 3, respectively. The first layer utilizes a stride of 2, reducing the feature resolution to half the original. The LiDAR depth image is then downsampled to half resolution and concatenated with the image feature map. This combined feature map is fed into the VoVNetV2 feature extractor, which operates through eight stages. The first four stages perform a 2× downsampling, while the subsequent stages upsample the feature map back to half resolution. 
At each up-sampling stage, the depth image is resized and concatenated with the feature map before being fed into the next stage. Finally, the feature maps are passed through a 1 × 1 convolution to decode the outputs. For object category (user-defined channels), 2D center (2 channels), 2D size (2 channels), 3D centroid (3 channels), 3D size (3 channels), range (1 channel), and orientation (1 channel), each output heatmap is of size (H/2 × W/2) × individual channels. Additionally, the head produces the LiDAR miscalibration output, which consists of three scalar values representing pitch, yaw, and roll in degrees.

\begin{figure*}
    \centering
    \includegraphics[width=0.75\linewidth]{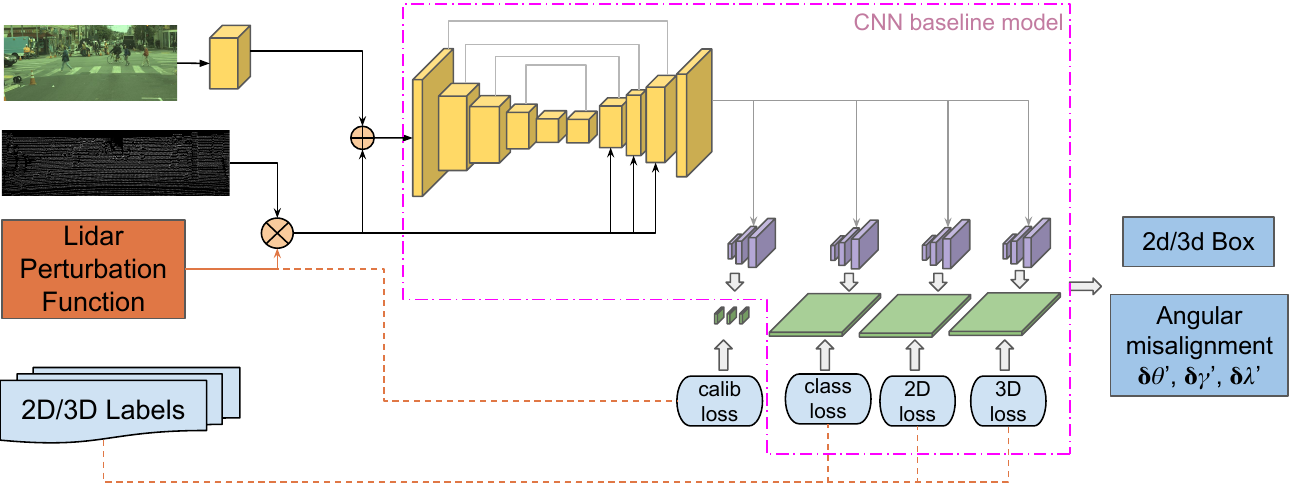}
    \caption{Our proposed multi-task network architecture and data augmentation approach. The network takes in RGB image and LiDAR points projected onto a virtual image. During the training, projected LiDAR points are perturbed in a controlled fashion, and the network is tasked to predict the parameters of that perturbation along with detecting and classifying 3D objects in the scene. This synthetic data augmentation and training technique ensures the model is robust to small perturbations in real-world scenarios.}
    \label{fig:enter-label}
\end{figure*}

\subsubsection{Loss Functions}
Below, we iterate over all the loss functions used to train the multi-task network mentioned in the previous section. We apply different weights $W_{\{c, o, s, \hat{o}, \hat{s}, d, \phi, \theta\}}$ on each loss for multi-task learning.

First, we use focal loss\cite{focalloss} to supervise the category output:
\begin{equation}
\mathcal{L}_{\text{class}} = -\frac{W_c}{N} \sum_{i}^{N} \alpha_i (1 - p_i)^{\gamma} \log(p_i)
\end{equation}
where $N$ denotes the number of pixels, $p_i$ represents the predicted probability for the ground-truth class at pixel i, and $(\alpha,\gamma)$ are focal loss weight parameters. 

In addition to classification loss, for reliable 2D detection location and extent prediction; we supervise the offsets ($o_{i}^{x}, o^{y}_{i}$) of the heatmap pixel, the width $w^{\text{2D}}_{j}$ and the height $h^{\text{2D}}_{j}$ of each object bounding box $j$. To handle uncertainties, the network predicts both the mean and the diversity of a Laplacian distribution. The loss function is applied to minimize the negative log-likelihood of the Laplacian distribution \cite{LLD}. The diversity parameter $b$ is used as the uncertainty value \cite{nair2022maximum}.   

Here $^*$ denotes the model's predictions, and $b_{\{ o^{x}_{i}, o^{y}_{i}, w_i^{\text{2D}}, h_i^{\text{2D}} \}}$ is the diversity parameter for each distribution \cite{LLD}.

\begin{equation}
\begin{aligned}
\mathcal{L}_{\text{2D}} &= \frac{1}{N} \sum_{i=1}^{N} \Bigg( W_o ( \frac{\| o^{x}_{i} - o^{x*}_{i} \|_1}{b_{o^{x}_{i}}} + \frac{\| o^{y}_{i} - o^{y*}_{i} \|_1}{b_{o^{y}_{i}}})  \\ 
&\quad + W_s (\frac{\| w^{\text{2D}}_{i} - w_{i}^{\text{2D}*} \|_1}{b_{w_i^{\text{2D}}}} + \frac{\| h^{\text{2D}}_{i} - h_{i}^{\text{2D}*} \|_1}{b_{h_i^{\text{2D}}}}) \\ 
&\quad + W_o \log(b_{o^{x}_{i}}b_{o^{y}_{i}}) + W_s  \log(b_{w_i^{\text{2D}}}b_{h_i^{\text{2D}}}) \Bigg) 
\end{aligned}
\end{equation}

Similarly, to accurately estimate the position of the detected object in 3D, the losses $\mathcal{L}_{\text{3D}}$ for the 3D position and extent parameters are computed in an analogous way. We compute the offsets ($\hat{o}_{i}^{x}, \hat{o}^{y}_{i}$) of the heatmap pixel by projecting the 3D box centroid onto 2D. We use L1 loss to supervise the 3D Euclidean distance $d_i$ and the orientation $\phi_i$.

\begin{equation}
\begin{aligned}
\mathcal{L}_{\text{3D}} &= \frac{1}{N} \sum_{i=1}^{N} \left( W_{\hat{o}}(\frac{\| \hat{o}^{x}_{i} - \hat{o}^{x*}_{i} \|_1}{b_{\hat{o}^{x}_{i}}} + \frac{\| \hat{o}^{y}_{i} - \hat{o}^{y*}_{i} \|_1}{b_{\hat{o}^{y}_{i}}}) \right. \\ 
& + \left. W_{\hat{s}}(\frac{\| w^{\text{3D}}_{i} - w_{i}^{\text{3D}*} \|_1}{b_{w_i^{\text{3D}}}} + \frac{\| l^{\text{3D}}_{i} - l_{i}^{\text{3D}*} \|_1}{b_{l_i^{\text{3D}}}} + \frac{\| h^{\text{3D}}_{i} - h_{i}^{\text{3D}*} \|_1}{b_{h_i^{\text{3D}}}}) \right. \\  
& + \left. W_{\hat{o}}\log(b_{\hat{o}^{x}_{i}}b_{\hat{o}^{y}_{i}}) + W_{\hat{s}}\log(b_{w_i^{\text{3D}}}b_{h_i^{\text{3D}}}b_{l_i^{\text{3D}}}) \right. \\ 
& + W_d \| d_{i} - d_{i}^* \|_1 + W_\phi \| \phi_{i} - \phi_{i}^* \|_1 \Bigg)
\end{aligned}
\end{equation}

The miscalibration loss is supervised in pitch, yaw, and roll axes as shown in equation \ref{eq:mis-calib}:

\begin{equation}
\label{eq:mis-calib}
\begin{split}
\mathcal{L}_{\text{miscal}} &= W_\theta \Bigg( \frac{\| \theta^{\text{pitch}} - \theta^{\text{pitch*}} \|_1}{b_{\theta^{\text{pitch}}}} 
+ \frac{\| \theta^{\text{yaw}} - \theta^{\text{yaw*}} \|_1}{b_{\theta^{\text{yaw}}}} \\
&\quad + \frac{\| \theta^{\text{roll}} - \theta^{\text{roll*}} \|_1}{b_{\theta^{\text{roll}}}} 
+ \log(b_{\theta^{\text{pitch}}} b_{\theta^{\text{yaw}}} b_{\theta^{\text{roll}}}) \Bigg)
\end{split}
\end{equation}

Finally, the total loss is formed as 
\begin{equation}
    \mathcal{L}_{\text{total}} = \mathcal{L}_{\text{class}} + \mathcal{L}_{\text{2D}} + \mathcal{L}_{\text{3D}} + \mathcal{L}_{\text{miscal}}
\end{equation}
 We choose multi-task weights $W_o$ = $1.0, W_c$ = $2.0, W_s$ = $0.1, W_{\hat{o}}$ = $0.25, W_{\hat{s}}$ = $1.0, W_d$ = $1.5, W_\phi$ = $0.1,$ and $W_\theta$ = $0.4$ for the experiments.

\subsubsection{Training Details}
To ensure that our network is robust against sensor misalignment and also predicts LiDAR misalignment accurately, we apply LiDAR perturbation (fault injection) during training. We use Gaussian distribution with a standard deviation of 0.5 degrees to perturb the rotation of LiDAR points within the camera frame along each axis of rotations. These perturbations affect the pitch, yaw, and roll axes up to a maximum of 1.0 degree.

Our method is trained on inputs comprising a single image and 100ms of LiDAR data centered on the image timestamp. Training is performed for 450k iterations using an Adam optimizer\cite{kingma2014adam} with an initial learning rate of 8e-4, decaying by 0.9 every 4000 iterations. We leverage NVIDIA A10 GPUs on AWS g5.4xlarge instances for training.

\subsection{Phase 3: Post-Processing}
\label{sec:post}
For the final phase of our system, we process the output from the network to obtain 3D detection and accurate misalignment estimation. 

\subsubsection{3D Detections} To leverage the outputs of our network, which include per-pixel classification, 2D object detection, and 3D object detection heads, we employ a 3D version non-maximum suppression (NMS) strategy similar to CenterNet \cite{centernet}. 


\subsubsection{Misalignment Fusion} 
We leverage the predicted uncertainty (diversity parameter $b_{\theta^{\text{pitch}}}$, $b_{\theta^{\text{yaw}}}$, and $b_{\theta^{\text{roll}}}$) values from network to fuse the misalignment estimates over a 5-second window. First, we filter out all the predictions with an uncertainty value higher than $0.3$ degree. Then, we perform a weighted average over the remaining measurements.



\subsubsection{Self Correction}

At module level, we use the fused prediction and apply a correction to incoming LiDAR data to align with camera image input by updating the extrinsic parameters.  
\begin{equation}
\label{eq:corr}
\begin{aligned}
T_{corrected} = T \cdot [R_{estimate} | 0] 
\end{aligned}
\end{equation}

where $R_{estimate}$ is the rotational matrix corresponding to the fused $[\theta^{roll}, \theta^{pitch}, \theta^{yaw}]$ predictions and translation component is set to 0. During the future iteration $T_{corrected}$ is used in the phase-1 for 3D point projection onto virtual image $I_{depth}$. 
In section \ref{sec:result}, we demonstrate that feeding the model corrected data improves the model performance on critical object detection tasks.

\section{Verification}

We employ a fault injection methodology that simulates misalignment conditions by perturbing the extrinsic parameters on a carefully curated test dataset. Specifically, we introduce controlled misalignment by altering the extrinsic parameters of the LiDAR sensor, which defines its spatial orientation and position relative to the vehicle's coordinate frame. These faults are injected by perturbing the roll, pitch, and yaw angles within a range of -1.0 to 1.0 degrees of rotation. This range is selected to emulate realistic scenarios where minor misalignments can occur due to factors such as vibrations, impacts, or thermal effects. In the experiment section, we show such minor misalignment is enough to have a severe impact on long range perception (45\% regression in 300-400 meters from Table \ref{table:detection_perf}).

By systematically injecting these perturbations across the test dataset, we can comprehensively evaluate the performance of our proposed method in detecting and quantifying the misalignment, as well as assess the effectiveness of the mitigation strategies, such as self-correction of extrinsic parameters. 

\begin{figure}[h]
\centering
\begin{tikzpicture}
\begin{axis}[
    xlabel={Lidar Misalignment Injection in Degrees},
    ylabel={Prediction Error in Degrees},
    xmin=-1.0,
    xmax=1.0,
    ymin=0.,
    ymax=0.08,
    grid=both,
    legend style={font=\scriptsize, at={(0.5, 0.15)}, anchor=north,legend columns=3},
    width=\columnwidth,
    height=6cm,
    tick label style={font=\scriptsize},
    label style={font=\scriptsize},
]

\addplot [mark=dot, blue, error bars/.cd, y dir=both, y explicit] coordinates {
    (-0.9, 0.06149857600580379) +- (0.008082929721253648, 0.008082929721253648)
    (-0.7, 0.054504944783725716) +- (0.008248712353836729, 0.008248712353836729)
    (-0.49999999999999994, 0.050343891657134446) +- (0.00680739035930834, 0.00680739035930834)
    (-0.29999999999999993, 0.05837739806719385) +- (0.009867579678237991, 0.009867579678237991)
    (-0.09999999999999998, 0.05363370175024623) +- (0.008316065104609204, 0.008316065104609204)
    (0.10000000000000009, 0.05243451625788555) +- (0.007232660848073509, 0.007232660848073509)
    (0.30000000000000016, 0.05469563959986593) +- (0.007875557240637645, 0.007875557240637645)
    (0.5000000000000001, 0.0541054386275387) +- (0.006495416456089824, 0.006495416456089824)
    (0.7000000000000001, 0.05121636933194002) +- (0.004551573884643744, 0.004551573884643744)
    (0.9, 0.05167044929559903) +- (0.003763976737387649, 0.003763976737387649)
};

\addplot [mark=dot, red, error bars/.cd, y dir=both, y explicit] coordinates {
    (-0.9, 0.02590229946838282) +- (0.0006680427227756074, 0.0006680427227756074)
    (-0.7, 0.02958762101408258) +- (0.0015509980519294458, 0.0015509980519294458)
    (-0.49999999999999994, 0.030562440575718913) +- (0.0018133175498585934, 0.0018133175498585934)
    (-0.29999999999999993, 0.03104448726929304) +- (0.0017226620155799176, 0.0017226620155799176)
    (-0.09999999999999998, 0.030728672855281847) +- (0.0014892287694072248, 0.0014892287694072248)
    (0.10000000000000009, 0.030002818525822712) +- (0.0018135629555934318, 0.0018135629555934318)
    (0.30000000000000016, 0.02982765765269583) +- (0.0012585374461975903, 0.0012585374461975903)
    (0.5000000000000001, 0.03408560367024236) +- (0.0017003389778514494, 0.0017003389778514494)
    (0.7000000000000001, 0.03688480586208903) +- (0.0017291536306757145, 0.0017291536306757145)
    (0.9, 0.03979090708408612) +- (0.0015027037507302076, 0.0015027037507302076)
};

\addplot [mark=dot, green, error bars/.cd, y dir=both, y explicit] coordinates {
    (-0.9, 0.017461581293217524) +- (0.0003896069273143294, 0.0003896069273143294)
    (-0.7, 0.020856916687354843) +- (0.0006351167938949153, 0.0006351167938949153)
    (-0.49999999999999994, 0.018642300242534163) +- (0.00038986653874655274, 0.00038986653874655274)
    (-0.29999999999999993, 0.019023562286308256) +- (0.000500185218323029, 0.000500185218323029)
    (-0.09999999999999998, 0.019173225886116474) +- (0.0006176566432087798, 0.0006176566432087798)
    (0.10000000000000009, 0.01837949136555876) +- (0.00042357025286776873, 0.00042357025286776873)
    (0.30000000000000016, 0.0173979975922877) +- (0.00037806463724224796, 0.00037806463724224796)
    (0.5000000000000001, 0.02001910531698809) +- (0.00044465718339804, 0.00044465718339804)
    (0.7000000000000001, 0.02265508749357136) +- (0.0004334434761682759, 0.0004334434761682759)
    (0.9, 0.022962262354167606) +- (0.00041779892489294434, 0.00041779892489294434)
};

\legend{roll, pitch, yaw}
\end{axis}
\end{tikzpicture}
\caption{Mean and variance observed in misalignment estimation error at different levels of perturbations for each rotational dimension.}
\label{fig:mean-sigma-error}
\end{figure}
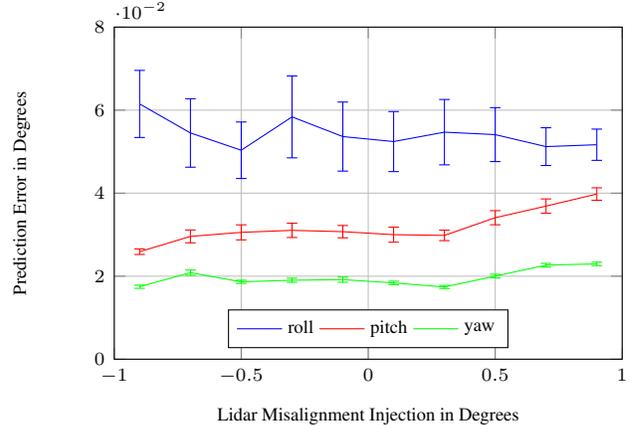

\subsection{Experimental Setting}

\subsubsection{Dataset}
To validate our proposed approach, we leverage our diverse internal dataset. Our focus in this study is mainly the degradation in long-range detections due to small misalignments. This limits us from using publicly available datasets such as Waymo Dataset which provides 3D labels only until 85 meters range \cite{sun2020scalability} and KITTI \cite{Geiger2013IJRR} around 70 meters. To provide better reproducibility and to show applicability of this approach for short range perception, we also conduct experiments on Waymo dataset and compare the results against available approaches. The limitations will be further discussed in the following sections.

Our internal proprietary dataset provides 3D bounding box labels up to 500m range using Frequency Modulated Continuous Wave (FMCW) LiDAR system. It comprises 43,500 five-second snippets for training and 1,000 for validation, featuring synchronized 8MP camera images with a 30° field of view, LiDAR data from FMCW LiDAR system. More importantly, this dataset enables us to verify our method's effectiveness on distant objects (upto 500 meter range) under diverse conditions.
From the dataset, we extract 5-second long continuous snippets of sensor data stream. To simulate realistic misalignment scenarios, we employ purpose built fault injector tool to perturb the incoming LiDAR point data within each snippet, introducing roll, pitch and yaw perturbations uniformly sampled in the range of [-1 degree to 1 degree] with an interval of 0.1 degree with total 1000 snippets of data.

\begin{table}[h]
\centering
\captionsetup{font=small}
\renewcommand{\arraystretch}{1.5} 
\scriptsize 
\begin{tabularx}{\columnwidth}{|>{\centering\arraybackslash}m{1.7cm}|>{\centering\arraybackslash}m{0.8cm}|>{\centering\arraybackslash}m{0.8cm}|>{\centering\arraybackslash}m{0.8cm}|>{\centering\arraybackslash}m{0.8cm}|>{\centering\arraybackslash}m{0.8cm}|}
\hline
\multicolumn{6}{|c|}{\textbf{On Internal Proprietary Long Range Dataset with $\delta \in [-1^\circ, 1^\circ]$}} \\
\hline
\multirow{2}{*}{\textbf{Fusion Approach}} & \multicolumn{2}{c|}{\textbf{Misalig. Acc.}} & \multicolumn{3}{c|}{\textbf{Mean Est. Error}} \\
\cline{2-6}
 & \textbf{Prec. $\uparrow$} & \textbf{Rec. $\uparrow$} & \textbf{Roll $\downarrow$} & \textbf{Pitch $\downarrow$} & \textbf{Yaw $\downarrow$} \\
\hline
Per Frame & 0.9813 & 0.9637 & 0.0700$^\circ$ $\pm 0.017$  & 0.0313$^\circ$ $\pm 0.008$ & 0.0241$^\circ$ $\pm 0.006$\\
\hline
Snippet (w/o uncertainty) & 0.9821 & 0.9644 & 0.0620$^\circ$ $\pm 0.015$ & 0.0298$^\circ$ $\pm 0.005$ & 0.0184$^\circ$ $\pm 0.006$ \\
\hline
Snippet (w uncertainty) & 0.9861 & 0.9650 & 0.0450$^\circ$ $\pm 0.008$ & 0.0290$^\circ$ $\pm 0.001$ & 0.0141$^\circ$ $\pm 0.0007$ \\
\hline
\end{tabularx}



\caption{Performance metrics with CenterNet baseline \cite{centernet} per frame, per snippet with fused predictions without considering uncertainty, and per snippet with fused predictions considering uncertainty (filtering out estimates with high predicted uncertainty and performing weighted average over remaining frames).}
\normalsize 
\label{table:miscalib_pred}
\end{table}

\subsubsection{Evaluation Metrics} Here we describe different metrics used to demonstrate the effectiveness of our approach. For the internal long-range dataset, we apply a 1.0 degree LiDAR perturbation, as previously described. For the Waymo dataset, a 5.0 degree perturbation is applied to create more pronounced misalignment in short-range scenarios.
\begin{table}[h]
\centering
\scriptsize
\renewcommand{\arraystretch}{1.5} 
\begin{tabularx}{\columnwidth}{|>{\centering\arraybackslash}X|>{\centering\arraybackslash}X|>{\centering\arraybackslash}X|>{\centering\arraybackslash}X|}
    \hline
    \textbf{Method} & \textbf{Pitch°} & \textbf{Yaw°} & \textbf{Roll°} \\
    \hline
    Yuan \cite{Yuan} & 0.411 & 0.325 & 0.423 \\
    \hline
    Zhu \cite{Zhu} & 0.421 & 0.301 & 0.305 \\
    \hline
    Sun \cite{Sun} & \textbf{0.202} & 0.211 & \textbf{0.171} \\
    \hline
    \textbf{Ours} & 0.248 & \textbf{0.068} & 0.309 \\
    \hline
\end{tabularx}
\caption{Comparison of different methods for pitch, yaw, and roll error on Waymo dataset \cite{sun2020scalability}.}
\label{table:methods}
\end{table}

\textbf{Misalignment Detection Accuracy:} 
The misalignment detection accuracy metric (see MDA columns in Table \ref{table:miscalib_pred}) evaluates our method's ability to reliably identify sensor alignment degradation while minimizing false alarms. We evaluate results as \textbf{positive} when our method predicts misalignment of $>0.1$ degree and \textbf{negative} when our method predicts misalignment of $<0.1$ degree. 
\textbf{True-Positive} is considered, when both predicted and injected misalignment are $>0.1$ degree and \textbf{True-Negative} is considered when both predicted and injected misalignment are $<0.1$ degrees. Using this, we report recall and precision across all snippets under varied conditions to comprehensively assess misalignment detection performance.  

\textbf{Misalignment Estimation Error Analysis:}
We evaluate the misalignment estimation error at both the per-frame and per-snippet levels. For per-frame evaluation, we directly compare the predicted misalignment values against the injected ground truth misalignment. For per-snippet evaluation, we compute the mean of all frame-level misalignment predictions within a snippet. Additionally, we leverage the predicted uncertainty estimates to fuse the outputs by ignoring predictions with high uncertainty (sigma $> 0.3$ degrees) and computing a weighted average over the remaining predictions as described in section \ref{sec:post}.

\textbf{Misalignment Estimation Error Benchmark:} 
To better evaluate the effectiveness of our approach, we compare it with other target-less learning methods on the Waymo dataset \cite{sun2020scalability}. We leverage the experiment settings from Sun et al. \cite{Sun}, where perturbations are applied differently at each group, and the mean average error is calculated as the final result. 
However, we compute the average over 30,000 frames, as opposed to the 60 frames in the original setting  \cite{Sun}, since we observed result fluctuations across different groups.  The comparison is presented in Table \ref{table:methods}. 

\textbf{Perception Performance Metrics:}
In addition to CenterNet, we include SpotNet \cite{foucard2024spotnet} in this experiment to demonstrate the effectiveness of the proposed method as an auxiliary task. SpotNet is a long range, image-centric, and LiDAR-anchored approach that jointly learns 2D and 3D detection tasks.  We measure the 3D detection performance of our multi-task model against a baseline and ensure that our module-level mitigation method strictly increases perception performance across all conditions (see Table \ref{table:detection_perf}). We employ the max F1 score metric (IoU=0.1) to evaluate the Bird's Eye View (BEV) object detection performance for actors at different ranges under various fault injection scenarios. As a baseline, we show the result from the model that is not trained with any perturbation. We compare the results of our proposed network architecture with and without applying the correction. This comprehensive evaluation helps us demonstrate the tangible benefits of our mitigation strategy in improving perception robustness and maintaining reliable operation in the presence of sensor misalignment faults. 

Additionally, for the Waymo dataset, we include only the results from SpotNet due to a label association issue. In Waymo, the 2D and 3D labels are not linked, preventing us from directly associating 2D labels with their corresponding 3D labels. As a result, we are unable to train CenterNet on 3D detection, since our method relies on 2D labels to encode target image pixels and then trains the 3D attributes from the associated 3D labels. Therefore, we compare only the 2D detection task on the Waymo dataset (see Table \ref{table:waymo_performance}).

\begin{tiny}

\begin{table*}[h]
\centering
\resizebox{0.8\textwidth}{!}{ 

\begin{tabularx}{\textwidth}{|c|>{\centering\arraybackslash}p{1.45cm}|>{\centering\arraybackslash}p{1.45cm}|>{\centering\arraybackslash}p{2.45cm}|>{\centering\arraybackslash}p{2.45cm}|>{\centering\arraybackslash}p{2.5cm}|>{\centering\arraybackslash}p{2.5cm}|}

\hline
 \multicolumn{7}{|c|}{\textbf{BEV Max-F1 @ 0.1 IoU for Vehicles with CenterNet \cite{centernet} $\uparrow$}} \\
\hline
\textbf{Range [m]} & \multicolumn{2}{c|}{\makecell{\textbf{Baseline}}} & \multicolumn{2}{c|}{\makecell{\textbf{Proposed} \\ \textbf{(No Correction Applied)}}} & \multicolumn{2}{c|}{\makecell{\textbf{Proposed} \\ \textbf{(Predicted Correction Applied)}}} \\
\cline{2-7}
& \textbf{calib} & \textbf{miscal} & \textbf{calib} & \textbf{miscal} & \textbf{calib} & \textbf{miscal} \\
\hline

200-300 & 0.177 & 0.104 & \textbf{0.250 (+41.2\%)} & 0.249 (+39.4\%) & \textbf{0.250 (+41.2\%)}  & \textbf{0.251 (+41.3\%)} \\
\hline
300-400 & 0.111 & 0.061 & 0.168 (+51.3\%) & 0.169  (+77.0\%) & \textbf{0.170 (+53.13\%)} & \textbf{0.170 (+78.6\%)} \\
\hline
400-500 & 0.082 & 0.075 & 0.156 (+90.2\%) & 0.156 (+90.2\%) & \textbf{0.161 (+96.3\%)} & \textbf{0.159 (+112\%)} \\
\hline

\multicolumn{7}{|c|}{\textbf{*BEV Max-F1 @ 0.1 IoU for Vehicles with SpotNet \cite{foucard2024spotnet} $\uparrow$}} \\ 
\hline
\textbf{Range [m]} & \multicolumn{2}{c|}{\makecell{\textbf{Baseline}}} & \multicolumn{2}{c|}{\makecell{\textbf{Proposed} \\ \textbf{(No Correction Applied)}}} & \multicolumn{2}{c|}{\makecell{\textbf{Proposed} \\ \textbf{(Predicted Correction Applied)}}} \\
\cline{2-7}
& \textbf{calib} & \textbf{miscal} & \textbf{calib} & \textbf{miscal} & \textbf{calib} & \textbf{miscal} \\
\hline
200-300 & \textbf{0.804} & 0.360 & 0.794 (-1.2\%) & 0.789 (+119\%) & 0.789 (-1.8\%)  & \textbf{0.792 (+124.3\%)} \\
\hline
300-400 & \textbf{0.734} & 0.250 & 0.726 (-1.0\%) & 0.718  (+187\%) & 0.726 (-1.0\%) & \textbf{0.721 (+188\%)} \\
\hline
400-500 & \textbf{0.640} & 0.220 & 0.620 (-3.12\%) & 0.601 (+173\%) & 0.626 (-3.06\%) & \textbf{0.626 (+184\%)} \\
\hline
\end{tabularx}
}
\captionsetup{font=small}
\caption{Comparison of 3D vehicle detection performance.  The "Proposed (No Correction Applied)" column represents the performance of our multi-task model when evaluated on miscalibrated (\textbf{miscal}) and calibrated (\textbf{calib}) data without applying any correction. The "Proposed (Predicted Correction Applied)" column shows the improved performance achieved by applying the extrinsic parameter correction using the misalignment values predicted by our model. For readers, we highlight the highest score in each row with bold characters for calibrated and miscalibrated test data category individually. \textit{*The dataset count slightly differs.}} 

\label{table:detection_perf}
\end{table*}



\end{tiny}

\subsection{Results}
\label{sec:result}
This section evaluates the performance of our proposed approach and its impact on enhancing safety across various driving scenarios. For system-level failure mitigation, we focus on the Misalignment Detection Accuracy (MDA) metrics, specifically precision and recall. As shown in Table \ref{table:miscalib_pred}, the results demonstrate the high precision and recall achieved by our method in detecting sensor misalignment across different conditions. Additionally, incorporating uncertainty estimates in the Per Snippet evaluation improves misalignment estimation accuracy, as evidenced by the lower Mean Estimated Error values.

Table \ref{table:methods}  presents results that are comparable to the benchmark in pitch and roll, while outperforming in yaw. By combining these table results with the misalignment estimates shown in the plot (Figure \ref{fig:mean-sigma-error}) of the injection LiDAR misalignment, we see that the model demonstrates promising performance in estimating yaw misalignment but struggles with roll misalignment. This behavior could be due to the differences in vertical and horizontal resolution for the LiDAR. LiDAR has a slightly higher horizontal resolution, which makes estimating yaw miscalibration easier compared to Pitch.  The roll noise is only visible at the very edge of the image, which typically consists of feature-less areas such as the sky or the ground surface closer to the camera (see Figure \ref{sec:intro}), making it more difficult for the model to observe feature displacement.


Table \ref{table:detection_perf} compares the performance of our multi-task model and a baseline model for vehicle detection on calibrated and miscalibrated data. We observe that the multi-task model significantly improves detection performance compared to the baseline when evaluated on the same miscalibrated data. 
It is important to note the significant performance drop in SpotNet \cite{foucard2024spotnet} under miscalibrated scenarios. This is because SpotNet is a LiDAR-anchored approach, making it highly sensitive to miscalibration. When there is misalignment between the projected sensor points and the image, the model mismatches the sensor data with the image background features, leading to decreased accuracy.
Furthermore, when applying correction (see qualitative example in Figure \ref{fig:cal-corr}) using the predicted misalignment values, the performance improves for ranges beyond 200 meters. Notably, the model already performs adequately without the calibration correction, as it is trained on miscalibrated data, making it inherently robust against misalignment errors and thus reducing the impact of the correction mechanism. Table \ref{table:waymo_performance} also shows that the proposed method improves short range 2D detection on the Waymo dataset

We also note that, even on the well-calibrated test set, the performance of the proposed model increases compared to the baseline; which was unexpected earlier. Upon closer investigation, we notice that in many instances our manually curated dataset has misalignment that went unnoticed. In a few cases, the alignment was accurate, and model improved in ultra long-range detections (400 meters and beyond), which further shows that LiDAR perturbation could be a powerful tool as augmentation for training.

\begin{table}[h]
\centering
\captionsetup{font=small}
\renewcommand{\arraystretch}{1.5} 
\scriptsize 

\begin{tabularx}{\columnwidth}
{|>{\centering\arraybackslash}m{3.4cm}|>{\centering\arraybackslash}m{1.8cm}|>{\centering\arraybackslash}m{1.8cm}|}
\cline{1-3}
\multicolumn{3}{|c|}{\textbf{On Short Range Waymo Dataset\cite{sun2020scalability} with SpotNet \cite{foucard2024spotnet}} with $\delta \in [-5^\circ, 5^\circ]$} \\
\cline{1-3}
{\textbf{Approach}} & {\textbf{Eval Data}} & {\textbf{Max-F1 for 2D Detection}}  \\
\cline{1-3}
\multirow{2}{*}{\textbf{Baseline}} & {{Calibrated}} & {{0.693}} \\
\cline{2-3}
 & MisCalibrated & 0.640   \\
\cline{1-3}
\multirow{2}{*}{\textbf{Proposed w/o input correction}} & {{Calibrated}} & {\textbf{0.712}} \\
\cline{2-3} 
 & MisCalibrated & \textbf{0.698}   \\
\cline{1-3}

\end{tabularx}

\caption{2D detection performance metrics shows improvement on short-range perception with our proposed augmentation learning approach.(note: For short-range waymo dataset, we inject $\delta \in [-5, 5]$ degree of miscalibration)}
\normalsize 
\label{table:waymo_performance}
\end{table}

\section{Conclusion}

In this research, we addressed the problem of making perception systems in autonomous vehicles robust against sensor misalignment. We proposed using a multi-task network trained with synthetic data augmentation to predict the degree of misalignment between LiDAR and camera sensors, alongside the detection task. We fuse the predictions using predicted data uncertainty and use the final output for self-correcting the input data. Finally, we evaluated the method on a large-scale, real-world dataset, perturbed with a purpose-built fault injection method, to demonstrate the effectiveness of our proposal. 

\section*{Acknowledgments}
We express our special thanks to Chi-Kuei Liu and Thuyen Ngo\footnote{For their help in developing the model and preparing the dataset.} for their invaluable contributions.

{\small
\bibliographystyle{ieee_fullname}
\bibliography{main}
}

\end{document}